\documentclass{article}
\usepackage{arxiv,times}

\wzkfinalcopy

\usepackage{amsmath,amsfonts,bm}

\def\eqref#1{equation~\ref{#1}}

\def\1{\bm{1}}

\DeclareMathAlphabet{\mathsfit}{\encodingdefault}{\sfdefault}{m}{sl}
\SetMathAlphabet{\mathsfit}{bold}{\encodingdefault}{\sfdefault}{bx}{n}

\usepackage{hyperref}
\usepackage{url}
\usepackage{amsmath}
\usepackage{booktabs}
\usepackage{array}
\usepackage{xspace}
\usepackage{graphicx}
\usepackage{enumitem}
\usepackage{xcolor}
\definecolor{mylightblue}{RGB}{0,102,204}

\definecolor{mylightblue}{RGB}{100,149,200}

\setlist[itemize]{itemsep=1pt,topsep=3pt,parsep=3pt,partopsep=1pt}
\newcommand{\benchname}{WorldCupArena\xspace}

\newcommand{\placeholderfigure}[2]{
  \fbox{\parbox[c][#1][c]{0.94\linewidth}{\centering\small #2}}}

\title{WorldCupArena: Fine-Grained Evaluation of Language Models and Deep-Research Agents on Football Forecasting}

\author{
Zhaokai Wang$^{1}$, Tianlin Gui$^{1}$, Jiayuan Rao$^{1}$, Shangzhe Di$^{1}$, Yihong Tang$^{2}$, Dingli Liang$^{3}$ \\
[1mm]    
~~~~~~~~$^1$Shanghai Jiao Tong University \quad $^2$McGill University \quad $^3$University College London\\
    [3mm]
    \qquad \qquad \qquad \textbf{~Github:}~~\textcolor{mylightblue}{\small \url{https://github.com/wzk1015/WorldCupArena/}} \\
}

\providecommand{\keywords}[1]{}
\providecommand{\Description}[1]{}

\renewcommand{\cite}[1]{\citep{#1}}

\begin{document}

\maketitle

\begin{abstract}

Predicting a football match before kickoff requires more than knowing past results: a model must use changing information and make a clear prediction before the answer is available. We present \benchname, a dynamic benchmark for language models and deep-research agents. The 2026 FIFA World Cup is its first evaluation, and the same process can be reused for future leagues and cups. Before each match, a model either receives a common evidence package or searches for information itself. It predicts the result and score, likely players and events, match statistics, and the outcome of the competition. After the match, these predictions are compared with the recorded result. We report result accuracy, exact-score accuracy, and a scoreline score that gives some credit when a predicted score is close but not exact, together with scores for the other prediction tasks. Across systems, similar result accuracy can mask larger differences in detailed predictions. Four systems predicted champion Spain, and two of them also recovered the exact final pairing. Compared with betting-market and human-fan baselines, the best system shows only small gains in result and exact-score accuracy, but a clearer gain in Scoreline. New schedules can be added as they begin, allowing the benchmark to evaluate future models without using outcomes that are already known. Code, predictions and evaluation scripts will be publicly released.

\end{abstract}


\section{Introduction}

Most language-model benchmarks ask questions whose answers are already known. They are useful for testing knowledge and reasoning, but they do not test whether a model can make a useful prediction before an event happens. Recent dynamic benchmarks update their questions, and forecasting benchmarks wait for future outcomes before assigning scores \cite{kasai2023realtimeqa,white2025livebench,jin2021forecastqa,karger2025forecastbench}. Most of them, however, still focus on a single answer or a coarse event outcome. They say little about whether a model can find current evidence and use it consistently across several related predictions.

\begin{figure*}[t]
  \centering

  \IfFileExists{figures/improved/figure-01.pdf}{\includegraphics[width=\linewidth]{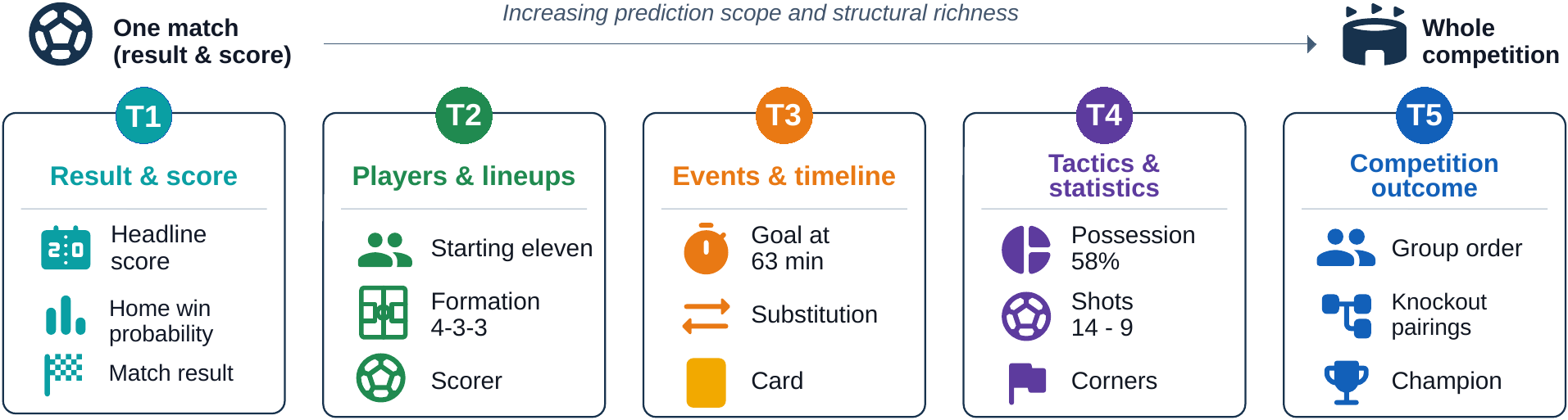}}{\placeholderfigure{1.32in}{Five evaluation layers from T1 result and score through T5 competition outcome, each with concrete prediction examples.}}

  \caption{Visual taxonomy of the five evaluation layers, moving from a match result and score through players and lineups, events, tactics and statistics, to the outcome of a whole competition.}
  \Description{Five connected modules showing examples for T1 result and score, T2 players and lineups, T3 events and timeline, T4 tactics and statistics, and T5 competition outcome.}
  \label{fig:layers}
\end{figure*}

Football provides a practical setting for this question.
Matches follow a public schedule, useful information changes before kickoff, and the final record contains much more than the winner. It also contains the score, lineups, goals and cards, match statistics, and the path through a league or cup. Existing sports benchmarks mainly study completed events \cite{deliege2021soccernetv2,rao2024matchtime,li2026sportsqa,xia2025sportu}, while traditional football forecasting usually focuses on the result or score distribution \cite{beal2021football,groll2019hybrid}. Evaluating all of these outputs together shows whether a plausible match prediction is supported by plausible details (Figure~\ref{fig:layers}).

\benchname evaluates models in this setting.
Twenty-four hours before a match, each model either receives the same evidence package or searches for its own sources.
It then submits probabilities, a predicted score, detailed match predictions, and a written explanation.
Once the match is over, the prediction is scored against the official record.
The same procedure can be applied to a new league or cup, so the benchmark can continue to test models released after the World Cup.

This paper reports 13 systems on the 2026 World Cup.
The main finding is that result accuracy alone hides important differences.
Models that predict a similar fraction of winners correctly can differ on scorelines, players, events, and statistics.
We also find no reliable gain from adding search to match prediction, and several models fail on the same upsets.
Against betting-market and human-fan baselines, the best model improves strict result prediction only slightly, while its larger Scoreline gain comes from making closer score predictions when it misses.

Our contributions are threefold:

\begin{itemize}[leftmargin=*]

  \item We build a benchmark that can be reused for future football competitions and report its first evaluation on the 2026 World Cup.
  \item We compare models given the same evidence with agents that search for evidence themselves, while asking both groups to make the same match and competition predictions.
  \item We score exact answers and close misses separately and publish the predictions and scoring code needed to check the results.
\end{itemize}

\section{Related Work}

\textbf{LLMs and deep-research agents.}
Language-agent benchmarks test browsing, planning, and long research tasks \cite{yao2023react,mialon2024gaia,yoran2024assistantbench,bosse2025deepresearchbench}.
We instead ask whether search improves a forecast about a match that has not happened. We compare models given common evidence (S1) with agents that gather their own evidence (S2), then score both after the match.

\noindent\textbf{Sports understanding and forecasting.}
SoccerNet-v2, MatchTime, UniSoccer, Sports-QA, and SPORTU evaluate video, commentary, and sports reasoning after a game has been played \cite{deliege2021soccernetv2,rao2024matchtime,rao2025unisoccer,li2026sportsqa,xia2025sportu}.
Football forecasting studies typically target results or score distributions, using statistics, news, or tournament simulations \cite{beal2021football,groll2019hybrid}. \benchname retains the before-match setting but evaluates a broader set of outputs, including players, events, statistics, and competition outcomes.

\noindent\textbf{Temporal prediction benchmarks.}
RealTime QA and LiveBench regularly add new questions so that their answers do not become stale \cite{kasai2023realtimeqa,white2025livebench}.
ForecastQA and Autocast use historical cut-off dates, whereas ForecastBench collects predictions before unknown outcomes \cite{jin2021forecastqa,zou2022autocast,karger2025forecastbench}.
\benchname follows the latter protocol: forecasts are locked before kickoff and scored afterward, with future leagues or cups supplying new matches.

\section{Benchmark Construction}

\begin{figure*}[t]
  \centering

  \IfFileExists{figures/improved/figure-02.pdf}{\includegraphics[width=\textwidth]{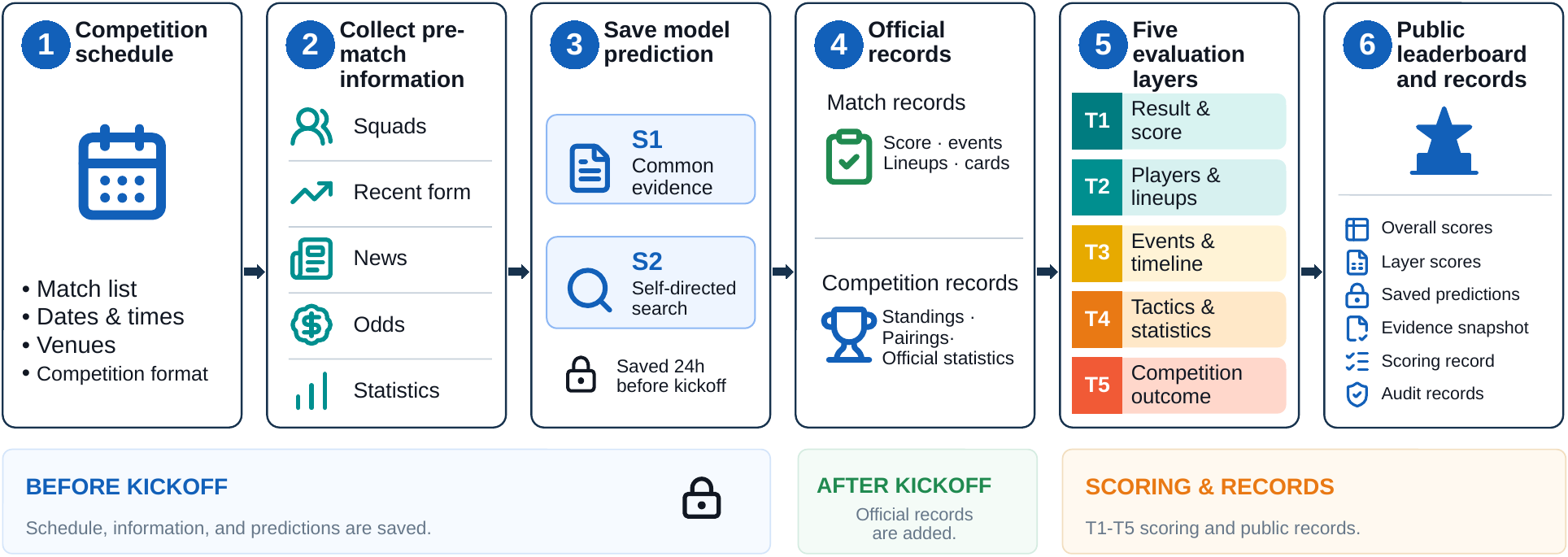}}{\placeholderfigure{1.30in}{A left-to-right benchmark lifecycle from competition scheduling and information collection through saved S1/S2 predictions, official records, five-layer scoring, and the public leaderboard.}}

  \caption{Overview of \benchname. Match information and model predictions are saved before kickoff. The final match and competition records are added afterward for scoring.}
  \Description{A six-stage diagram of the WorldCupArena prediction and evaluation lifecycle.}
  \label{fig:pipeline}
\end{figure*}

\benchname follows the same four steps for every match: register a future fixture, collect information until the prediction deadline, save each model's forecast, and score it after the official match record is available.
Applying the benchmark to another competition changes the schedule and competition rules, but not these four steps or the match-level metrics.
Figures~\ref{fig:layers} and~\ref{fig:pipeline} show what is predicted and how one match moves through the benchmark.

\subsection{Benchmark Scope and Protocol}

\noindent\textbf{What models predict.}
The benchmark asks for two kinds of prediction.
For each match, a model predicts the result and score, players, events, and match statistics.
It also predicts the competition as a whole.
Match predictions have the same form in any football competition.
Competition predictions need a small set of rules that describes what can be decided: for a league, this includes the final table, qualification, relegation, and champion.
For a cup, it includes group order, advancing teams, knockout pairings, and champion.
Only these competition rules change when a new league or cup is added.

This paper evaluates the 2026 World Cup.
Models first predict the group-stage fixtures.
The program then calculates each group table and the best third-place qualifiers from those predicted scores.
Finally, the models predict the knockout bracket produced by their own group forecasts \cite{fifa2026schedule}.
All competition-level forecasts are elicited and frozen before the tournament begins.
Later match outcomes are used only for scoring.
The same procedure can be configured before a future domestic league, continental cup, or World Cup begins.

\noindent\textbf{Evaluation layers.}
For brevity, we group the predictions into five layers: match result and score (T1), players and lineups (T2), match events (T3), tactics and statistics (T4), and the whole competition (T5). Table~\ref{tab:taxonomy} summarizes their targets, metrics, and weights. The next section explains how the layers are scored and combined.

\noindent\textbf{Temporal protocol.}
We save the evidence and prediction 24 hours before kickoff, so every system is judged using information available at the same time.
For agents that search the web, we also save the returned URLs and timestamps.
A prediction is excluded if it contains information published after the deadline or reveals the final result.

\subsection{Predictions and Ground Truth}
\noindent\textbf{Evidence settings.}
Each model returns the same JSON fields for result probabilities, predicted score, players, events, and match statistics, plus a written explanation. We test two ways of obtaining pre-match information.
In \textbf{S1}, every model receives the same package of squads, recent form, statistics, news, and available odds, and cannot use tools.
In \textbf{S2}, the agent receives only the fixture and must search for its own evidence.

\noindent\textbf{In-play records.}
For a subset of matches, we repeatedly give the current minute, score, events, and available match statistics to three models and ask them to update the final-result probabilities and predicted final score. Every checkpoint stores the observed match state and submission time. These in-play forecasts form a separate exploratory track and do not enter the main leaderboard, because they cover fewer models and their polling times are not perfectly regular.

\noindent\textbf{Forecast rationale.}
The written explanation covers the expected lineup, tactics, recent form and past meetings, important player matchups, injuries, and possible paths to a win, draw, or upset. It should also explain why the predicted result and score are consistent with the predicted players, events, and statistics.

\noindent\textbf{Truth acquisition and quality control.}
After the match, we collect the final score, lineups, events, and statistics independently of the model prediction.
Team and player identifiers are checked before scoring, and the normalized record is required to satisfy the basic score, event, and statistic consistency checks.

\begin{table*}[t]
  \caption{Evaluation taxonomy, metrics, and fixed layer and task weights.}
  \label{tab:taxonomy}
  \centering
  \scriptsize
  \renewcommand{\arraystretch}{1.2}
  \setlength{\tabcolsep}{2pt}
\vspace{2mm}
\resizebox{\textwidth}{!}{
  \begin{tabular}{
    >{\raggedright\arraybackslash}p{0.17\textwidth}
    >{\raggedright\arraybackslash}p{0.21\textwidth}
    >{\raggedright\arraybackslash}p{0.175\textwidth}
    >{\centering\arraybackslash}p{0.08\textwidth}
    >{\raggedright\arraybackslash}p{0.28\textwidth}}
    \toprule
    Layer & Representative targets & Metrics & Layer weight & Task weights \\
    \midrule
    T1 (Core result) & 1X2 probability, exact score, score closeness, goal difference & Brier, exact, scoreline score & 0.40 & 1X2 0.35, exact 0.20, scoreline 0.30, goal diff. 0.10, qualification 0.05 \\
    T2 (Player-level) & starters, formation, scorers, assists, player of match & Jaccard, F1, nDCG & 0.20 & starting eleven 0.30, formation 0.10, scorers 0.35, assists 0.15, player of match 0.10 \\
    T3 (Event-level) & goal time, substitutions, cards, penalties, own goals & matched MAE, event F1 & 0.15 & goal time 0.45, subs 0.20, cards 0.25, penalties 0.05, own goals 0.05 \\
    T4 (Tactics/statistics) & possession, shots, corners, passing, fouls, saves & sMAPE & 0.15 & formation 0.15, eight statistics 0.85 \\

    T5 (Competition-level) & group order, bracket, champion, scorers, awards & Kendall, bracket, top-1 & 0.10 & groups 0.25, bracket 0.35, champion 0.20, scorer 0.10, awards 0.10 \\
    \bottomrule
  \end{tabular}
}
\end{table*}

\section{Evaluation Method}

\subsection{Match-Level Metrics}
\noindent\textbf{Overview.}
All metrics are defined so that a larger value is better:

\begin{itemize}[leftmargin=*]

  \item \textbf{Composite} is the overall score that combines the five groups of predictions in Table~\ref{tab:taxonomy}, with more weight on the match result and score.
  \item \textbf{Result accuracy} is the percentage of matches for which the system correctly predicts home win, draw, or away win.
  \item \textbf{Exact-score accuracy} is the percentage for which it predicts both teams' goal totals exactly.
  \item \textbf{Scoreline score} also rewards close misses. For example, predicting 2--1 when the match ends 3--1 receives credit for the correct winner and a similar number of goals, but less than an exact prediction.
\end{itemize}

\noindent\textbf{Result and scoreline metrics.}
Result accuracy checks whether the most probable home/draw/away outcome is correct.
We also use the standard three-way Brier score to reward well-calibrated probabilities and penalize misplaced confidence \cite{brier1950verification}.
Exact-score accuracy is simply the percentage of matches for which both predicted goal totals equal the final score.

Exact accuracy treats every miss equally, so we additionally score how close the predicted score is.
An exact prediction receives 100. Otherwise, the score gives partial credit for the result class, goal difference, total goals, and team-wise goals.
For display, the mean Scoreline value uses a fixed monotonic calibration. 
Details are given in Appendix~\ref{app:formulas}.

\noindent\textbf{In-play aggregation.}
Each in-play checkpoint reuses result accuracy, the three-way Brier score, exact-score accuracy, and the raw Scoreline metric above.
We first average checkpoints within a match and then average across matches, so a match with more polls does not receive more weight.
To show how forecasts change during play, we report result accuracy for minutes 0--30, 31--60, 61--90, and 91 or later.
We also pair consecutive checkpoints whenever the observed score changes and compare the predictions immediately before and after that change.

\noindent\textbf{Player, event, and statistic metrics.}
For lineups and scorers, we measure how many predicted players appear in the match and reward a correct player more when the model ranks that player highly \cite{jarvelin2002cumulated}.
A predicted goal, card, or substitution is matched to the closest real event, with penalties for the wrong player or minute \cite{kuhn1955hungarian}.
Counts such as shots and corners are scored by percentage error using sMAPE \cite{hyndman2006accuracy}.

\noindent\textbf{Availability-aware aggregation.}
We use the term task-to-layer for the aggregation of task scores within T1--T5, and layer-to-composite for the aggregation of the five layer scores.
At both levels, the score is the fixed weighted mean over the target components defined in Table~\ref{tab:taxonomy}. 
The raw overall scores are converted with a fixed S-shaped transformation for display. 
Details are given in Appendix~\ref{app:formulas}.

\subsection{Competition-Level Evaluation}

A full-competition prediction evaluates group order, advancement, knockout pairings, and champion accuracy. The detailed notation and aggregation procedure are given in Appendix~\ref{app:formulas}.
The completed evaluation includes every group table, all knockout matchups, and champion Spain.
Top-scorer and award components are reported separately because their normalized competition-level targets are not part of this scoring release.
All competition-level predictions were elicited and frozen before the opening match.
No later result was supplied when the bracket or champion was submitted.

\section{Experiments}

\subsection{Models and Baselines}

We evaluate 13 systems: nine receive the prepared S1 evidence package, three are search-enabled S2 systems, and one is a deep-research agent. Every response is converted to the same prediction format before scoring.
The evaluated models are: Claude Opus 4.7 (Thinking) and Claude Opus 4.7 (Thinking + Search) \cite{anthropic2026claudeopus47}, GPT-5.4 and GPT-5.4 (Search) \cite{openai2026gpt54}, GLM-5.1 \cite{zai2026glm51}, Kimi K2.6 \cite{kimi2026k26}, MiniMax M2.7 \cite{minimax2026m27}, DeepSeek V4 Pro \cite{deepseek2026v4}, Gemini 3.1 Pro Preview (Thinking) and Gemini 3.1 Pro Preview (Thinking + Search) \cite{google2026gemini31pro}, Gemini Deep Research \cite{google2026deepresearch}, Doubao Seed 2.0 Lite \cite{bytedance2026seed2}, and Qwen3.7 Max \cite{qwen2026qwen37}.

The evaluated set is the complete 2026 World Cup schedule, covering the group stage, every knockout round, the third-place playoff, and the final \cite{fifa2026schedule}.
All systems use the same pre-match forecast contract.

Additionally, we consider three baselines:

\noindent\textbf{Polymarket baseline.} We take the last available home/draw/away price in the window $[T-24\mathrm{h},T)$, where $T$ is kickoff \cite{polymarket2026api}. 

\noindent\textbf{BetVictor baseline.} This baseline uses BetVictor's most likely match result and exact score, retrieved from Sportmonks~\cite{sportmonks2026odds}.
Sportmonks provides the latest stored pre-match odds rather than a guaranteed value from exactly 24 hours before kickoff, so this comparison is also descriptive rather than temporally matched. 

\noindent\textbf{Human-fan baseline.}
Human predictions come from an internal football-fan group. For each match, we use the most common predicted result and score. Tied modal answers share the credit equally. The archive records predictions made before kickoff but does not establish the same $T-24$ evidence horizon as S1, so this baseline is descriptive.

\begin{table*}[t]

  \caption{Results for the 2026 World Cup forecast track. Composite is the calibrated display value, Result and Exact are accuracies, and Scoreline is calibrated partial credit. S1 (standardized context) and S2 (self-directed search) denote the two evidence settings.}
  \label{tab:matchresults}
  \centering
  \small
  \renewcommand{\arraystretch}{1.2}
  \vspace{2mm}
\resizebox{0.95\textwidth}{!}{
  \begin{tabular}{llrrrr}
    \toprule
     Model & Setting & Composite & Result & Exact & Scoreline \\
    \midrule

     Claude Opus 4.7 (Thinking) & S1 & \textbf{33.76} & \underline{68.4} & \underline{15.8} & \underline{63.16} \\
     Claude Opus 4.7 (Thinking + Search) & S2 & \underline{31.26} & \textbf{70.7} & \textbf{17.2} & \textbf{68.49} \\
     GPT-5.4 & S1 & 29.91 & 65.7 & 11.8 & 56.05 \\

     GLM-5.1 & S1 & 28.59 & 68.0 & 13.6 & 62.27 \\
     GPT-5.4 (Search) & S2 & 28.14 & 68.0 & 10.3 & 58.78 \\
     Kimi K2.6 & S1 & 27.49 & 68.3 & 12.9 & 62.65 \\
     MiniMax M2.7 & S1 & 26.53 & 65.3 & 12.9 & 57.35 \\
     DeepSeek V4 Pro & S1 & 26.31 & 68.0 & 11.7 & 61.02 \\
     Gemini 3.1 Pro Preview (Thinking + Search) & S2 & 26.11 & 63.5 & 11.8 & 54.54 \\

     Gemini Deep Research & S2 & 25.49 & 58.6 & 12.1 & 41.36 \\
     Doubao Seed 2.0 Lite & S1 & 24.06 & 67.6 & 10.8 & 58.16 \\
     Gemini 3.1 Pro Preview (Thinking) & S1 & 23.69 & 59.8 & 12.7 & 36.95 \\
     Qwen3.7 Max & S1 & 22.78 & 61.5 & 10.6 & 39.85 \\
    \midrule
     Polymarket &  --  & -- & 65.4 & 8.7 & -- \\
     BetVictor & -- & -- & 68.3 & 16.3 & 53.35 \\
     Human fans & --  & -- & 69.7 & 15.6 & 53.40 \\
    \bottomrule
  \end{tabular}
}
\end{table*}

\begin{table*}[t]
  \caption{Scores of leading systems for result and score (T1), players and lineups (T2), events and timeline (T3), tactics and statistics (T4), and competition outcome (T5).}
  \label{tab:raw-layers}
  \centering
  \small
  \renewcommand{\arraystretch}{1.2}
  \vspace{2mm}
\resizebox{0.8\textwidth}{!}{
  \begin{tabular}{lrrrrr}
    \toprule
    Model & T1 & T2 & T3 & T4 & T5 \\
    \midrule
    Claude Opus 4.7 (Thinking) & 76.0 & 17.3 & 59.9 & 38.2 & 87.4 \\
    Claude Opus 4.7 (Thinking + Search) & 75.9 & 15.1 & 60.5 & 37.6 & 88.3 \\
    GPT-5.4 & 75.0 & 17.1 & 57.4 & 37.5 & 73.1 \\
    GLM-5.1 & 75.1 & 15.5 & 58.1 & 36.9 & 48.1 \\
    GPT-5.4 (Search) & 74.6 & 15.2 & 58.6 & 37.5 & 47.7 \\
    Kimi K2.6 & 75.5 & 15.1 & 56.2 & 36.6 & 47.5 \\
    MiniMax M2.7 & 74.6 & 14.2 & 58.6 & 36.4 & 47.9 \\
    DeepSeek V4 Pro & 74.8 & 15.5 & 55.1 & 36.8 & 49.0 \\
    Gemini 3.1 Pro Preview (Thinking + Search) & 74.1 & 15.3 & 56.8 & 36.9 & 47.0 \\
    \bottomrule
  \end{tabular}
}
\end{table*}

\begin{table*}[t]

  \caption{Full-competition predictions. Group measures predicted group order, Adv. measures whether the correct teams reach each round, Pair measures exact knockout matchups, Champ. is champion accuracy. T5 combines these four parts.}
  \label{tab:tournament}
  \centering
  \small
  \renewcommand{\arraystretch}{1.2}
  \vspace{2mm}
\resizebox{0.95\textwidth}{!}{
  \begin{tabular}{llrrrrr}
    \toprule
    Model & Setting & Group & Adv. & Pair & Champ. & T5 \\
    \midrule

    Claude Opus 4.7 (Thinking + Search) & S2 & \underline{87.5} & \textbf{87.0} & \textbf{71.0} & \textbf{100.0} & \textbf{88.3} \\
    Claude Opus 4.7 (Thinking) & S1 & 84.7 & \textbf{87.0} & \underline{70.4} & \textbf{100.0} & \underline{87.4} \\
    GPT-5.4 & S1 & 86.1 & 60.8 & 19.4 & \textbf{100.0} & 73.1 \\
    Doubao Seed 2.0 Lite & S1 & \textbf{88.9} & 58.7 & 15.3 & \textbf{100.0} & 72.8 \\
    DeepSeek V4 Pro & S1 & \textbf{88.9} & 60.7 & 20.2 & 0.0 & 49.0 \\
    GLM-5.1 & S1 & 86.1 & 60.9 & 19.2 & 0.0 & 48.1 \\

    MiniMax M2.7 & S1 & 86.1 & 60.4 & 19.0 & 0.0 & 47.9 \\
    GPT-5.4 (Search) & S2 & 84.7 & \underline{61.1} & 19.4 & 0.0 & 47.7 \\
    Kimi K2.6 & S1 & 84.7 & 60.7 & 18.3 & 0.0 & 47.5 \\
    Qwen3.7 Max & S1 & 84.7 & 59.9 & 18.3 & 0.0 & 47.2 \\

    Gemini 3.1 Pro Preview (Thinking + Search) & S2 & 86.1 & 59.1 & 15.3 & 0.0 & 47.0 \\
    Gemini Deep Research & S2 & \underline{87.5} & 35.3 & 19.2 & 0.0 & 40.7 \\
    Gemini 3.1 Pro Preview (Thinking) & S1 & 83.3 & 34.9 & 17.7 & 0.0 & 39.1 \\
    \bottomrule
  \end{tabular}
}
\end{table*}

\subsection{Match-Level Results}

Table~\ref{tab:matchresults} shows that result accuracy alone provides limited separation. Claude Opus 4.7 (Thinking) has the highest composite, while its search version leads result and exact-score accuracy. Exact scores remain difficult (10.3--17.2\%). Scoreline adds partial credit for near misses, and the raw layer values in Table~\ref{tab:raw-layers} explain the composite differences.

Claude Opus 4.7 (Thinking + Search) reaches 70.7\% result accuracy and 17.2\% exact-score accuracy, only modestly above the strongest human and market baselines. Its larger advantage is in Scoreline (68.49), indicating closer misses. These comparisons are descriptive because timestamps and fixture intersections differ.

Table~\ref{tab:raw-layers} shows that no system leads every layer: Claude Opus 4.7 (Thinking) leads T1--T2, its search version leads T3, and GPT-5.4 leads T4 among the systems shown. Similar result accuracy can therefore hide large differences in detailed predictions.

\begin{figure*}[t]
  \centering

  \IfFileExists{figures/improved/figure-03.pdf}{\includegraphics[width=\textwidth]{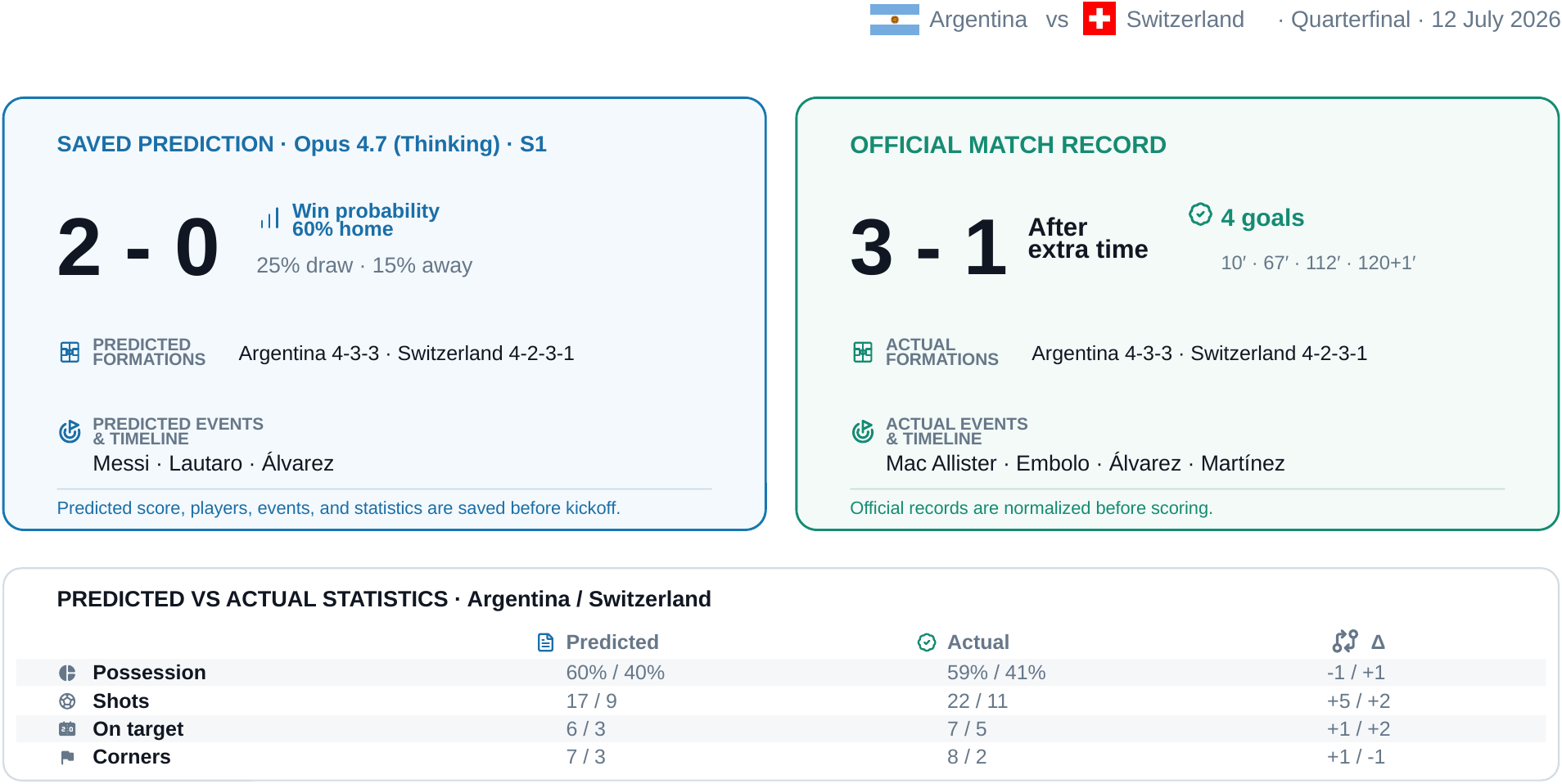}}{\placeholderfigure{2.02in}{A two-column case study: the left panel shows one saved prediction with score, probabilities, formations, players, events, and statistics, the right panel shows the corresponding official match record.}}

  \caption{Claude Opus 4.7 (Thinking)'s prediction for Argentina--Switzerland beside the recorded result. The example shows the predicted score, players, events, and statistics used in T1--T4.}
  \Description{Side-by-side view of one model prediction and the actual match outcome, including score, probabilities, formations, players, and match statistics.}
  \label{fig:case-study}
\end{figure*}

Figure~\ref{fig:case-study} shows how one saved prediction is compared with the match record.

On shared fixtures, the no-search version scores higher for all three paired systems. Search changes the displayed composite by $-0.26$ for Claude, $-4.26$ for GPT-5.4, and $-1.03$ for Gemini. Because these are complete products, the comparison is not a causal retrieval ablation.

All 13 systems selected France, Spain, Brazil, and Argentina as semifinalists, missing England. Four selected Spain as champion, and only the two Claude configurations also predicted the Spain--Argentina final.

\subsection{Competition-Level Results}

Table~\ref{tab:tournament} shows that group order is easier than the exact knockout route (83.3--88.9 versus 15.3--71.0). The Claude configurations lead T5 (88.3 and 87.4) because they predict both the Spain--Argentina final and champion Spain. Full-route scoring distinguishes a correct champion reached through the correct bracket from one reached through the wrong path.

\begin{table*}[t]
  \setlength{\tabcolsep}{3pt}
  \caption{In-play forecasting results. Metrics are averaged within each match and then across matches. Gap is the median wall-clock interval in minutes with the interquartile range in brackets. In panel (b), Before and After compare consecutive checkpoints whose observed score differs. A goal pair is precisely such a consecutive checkpoint pair.}

  \label{tab:live-results}
  \centering
  \renewcommand{\arraystretch}{1.2}
  \small
  \vspace{2mm}
  \resizebox{0.95\textwidth}{!}{
  \begin{tabular}{p{0.5\textwidth}crrrr}
    \toprule
    \multicolumn{6}{c}{\textbf{(a) Overall in-play performance}} \\
    Model & Gap & Result & Brier & Exact & Scoreline \\
    \midrule
    Claude Opus 4.7 (Thinking) & 7.0 [6.2, 13.4] & \textbf{70.5} & \textbf{79.8} & \textbf{30.7} & 76.9 \\
    GPT-5.4 & 6.7 [6.1, 11.8] & 64.3 & 78.5 & 27.4 & 72.8 \\
    Gemini 3.1 Pro Preview (Thinking) & 6.7 [6.2, 11.8] & 69.9 & 79.4 & 30.5 & \textbf{77.2} \\
    \bottomrule
  \end{tabular}
  }

  \vspace{3mm}
  \resizebox{\textwidth}{!}{
  \begin{tabular}{p{0.5\textwidth}rrrrrrr}
    \toprule
    \multicolumn{8}{c}{\textbf{(b) Result accuracy by match state}} \\
    Model & 0--30 & 31--60 & 61--90 & 91+ & Goal pairs & Before & After \\
    \midrule
    Claude Opus 4.7 (Thinking) & 66.1 & \textbf{64.4} & 75.1 & \textbf{80.2} & 208 & \textbf{66.9} & \textbf{78.9}  \\
    GPT-5.4 & 52.9 & 58.5 & 74.4 & 67.7 & 227 & 55.4 & 77.4  \\
    Gemini 3.1 Pro Preview (Thinking) & \textbf{68.3} & 64.1 & \textbf{75.8} & 73.8 & 229 & 64.6 & 78.5  \\
    \bottomrule
  \end{tabular}
  }
\end{table*}

\subsection{In-Play Results}

The in-play track uses the checkpoints that were actually recorded, not a fixed ten-minute schedule. The median wall-clock gap is 6.8 minutes (IQR 6.2--12.4), and the median change in provider-reported minute is 7 minutes (IQR 6--10).

Table~\ref{tab:live-results} shows similar overall performance with different strengths. Claude leads result, Brier, and exact-score accuracy, while Gemini has the highest raw Scoreline. All three perform best or near-best in minutes 61--90.

After an observed score change, result accuracy rises from 55.4--66.9\% to 77.4--78.9\%, and Brier scores improve by 8.4--12.1 points. These gains measure adaptation to new match state, not anticipation of the goal.

\subsection{Discussion}

\begin{table*}[t]
  \setlength{\tabcolsep}{3pt}

  \caption{Difficulty-stratified results: means over the eligible systems and displayed-composite leaders, Result and Exact are percentages.}
  \label{tab:difficulty}
  \centering
  \renewcommand{\arraystretch}{1.2}
  \footnotesize
   \vspace{2mm}
\resizebox{0.57\textwidth}{!}{
  \begin{tabular}{lrrrr}
    \toprule
    \multicolumn{5}{c}{\textbf{(a) Eligible-model means}} \\
    Slice & Comp. & Result & Exact & Score \\
    \midrule

    Balanced ($p_{\max}\leq0.50$) & 21.5 & 55.4 & 14.0 & 43.1 \\
    Heavy favorite ($p_{\max}\geq0.65$) & 29.2 & 75.5 & 9.6 & 70.3 \\
    Knockout & 34.9 & 70.7 & 18.4 & 72.2 \\
    Low scoring (outcome $\leq1$ goal) & 14.0 & 57.7 & 4.6 & 31.2 \\
    \bottomrule
    \vspace{1mm}
  \end{tabular}
} 
\resizebox{0.67\textwidth}{!}{
  \begin{tabular}{lp{0.5\textwidth}r}
    \toprule
    \multicolumn{3}{c}{\textbf{(b) Composite leaders}} \\
    Slice & Model & Comp. \\
    \midrule

    Balanced & Claude Opus 4.7 (Thinking) & 28.7 \\
    Heavy favorite & Claude Opus 4.7 (Thinking) & 38.2 \\
    Knockout & Claude Opus 4.7 (Thinking) & 46.1 \\
    Low scoring & Gemini 3.1 Pro Preview (Thinking + Search) & 17.9 \\
    \bottomrule
  \end{tabular}
}
\end{table*}

\noindent\textbf{Why we report several metrics.}
Result accuracy treats a cautious 45\% prediction and a confident 90\% prediction in the same way once both choose the same outcome.
The Brier score distinguishes them and penalizes misplaced confidence.
Exact-score accuracy answers a different question from the Scoreline metric, and the player, event, and statistic scores reveal different strengths again.
We therefore report all these values instead of asking one overall number to explain every aspect of a forecast.
The composite is a secondary summary of the fixed task hierarchy.
Scientific interpretation should begin with the raw per-layer values and the task-specific columns.

\noindent\textbf{Difficulty-stratified results.}
The main leaderboard averages different kinds of match.
Table~\ref{tab:difficulty} checks whether the result changes for four easier-to-understand groups:

\begin{itemize}[leftmargin=*]

  \item Balanced: matches in which the betting market gives no outcome more than 50\% probability after removing the bookmaker's margin.
  \item Heavy favorite: matches in which one outcome has at least 65\% probability.

  \item Knockout: elimination matches, including the third-place playoff and final.
  \item Low scoring: matches that finish with zero or one goal. This group is defined after the match and is used only to study a difficult pattern.
\end{itemize}

The first two groups use fixtures with valid pre-match odds.
Panel (a) shows the average system performance for each group.
Panel (b) shows the best individual system.
For example, 21.5 is the mean overall score on balanced matches, while 28.7 is the score of the best system on those matches.

The groups reveal different weaknesses.
Heavy favorites make the winner easier to predict but do not make the exact score easier.
Low-scoring matches are hardest overall: average result accuracy is 57.7\% and exact-score accuracy is 4.6\%.
Claude Opus 4.7 (Thinking) leads the first three groups, whereas Gemini 3.1 Pro Preview (Thinking + Search) leads the low-scoring group.

\noindent\textbf{Consensus-failure results.}
Table~\ref{tab:consensus-failures} lists matches in which every evaluated system chose the wrong result.
For Spain--Cape Verde, all systems selected Spain by 3--0 or 4--0, but the match ended 0--0 despite Spain's 89.1\% market probability.
In the third-place playoff, all systems selected France by a narrow margin, while England won 6--4.
These cases show that models can share the same favorite and fail together on unusual outcomes.

\begin{table*}[t]
  \caption{Consensus-failure case studies. Modal score counts are taken from the locked headline forecasts, Wrong 1X2 counts systems whose highest-probability result class disagrees with the final result.}
  \label{tab:consensus-failures}
  \centering
\renewcommand{\arraystretch}{1.2}
  \small
\vspace{2mm}
\resizebox{\textwidth}{!}{
  \begin{tabular}{lclp{0.49\textwidth}}
    \toprule
    Fixture & Actual & Predicted headline scores &  Failure pattern \\
    \midrule
    Spain--Cape Verde & 0--0 & $7\times$ 4--0, $6\times$ 3--0 & Extreme favorite held to a scoreless draw \\
    Netherlands--Japan & 2--2 & $12\times$ 2--1, $1\times$ 1--0 & Concentrated favorite-win and scoreline consensus \\

    France--England (3rd) & 4--6 & $6\times$ 2--1, $4\times$ 3--2, $3\times$ 3--1  & Unanimous favorite pick reversed in a ten-goal match \\
    \bottomrule
  \end{tabular}
}
\end{table*}

\noindent\textbf{Generalization beyond one tournament.}
This study is an empirical case study of one World Cup format.
The results should not be read as evidence that the same ranking or search behavior will hold in domestic leagues, less prominent competitions, or other sports.
Those settings differ in schedule density, roster stability, evidence availability, and competition rules.
The reusable part of \benchname is the pre-event lock, common output schema, objective truth interface, and metric contract.
Future tracks are needed to test transfer across those settings.

\section{Conclusion}

We introduced \benchname to evaluate pre-match football forecasts from language models and agents across results, scores, players, events, statistics, and competition outcomes.
Detailed metrics reveal differences that result accuracy alone misses, while search does not consistently improve match prediction and models can fail together on the same favorite.
Competition scoring and separate exact-score and Scoreline metrics distinguish correct routes and reasonable near misses.
The World Cup is the first track; future leagues, cups, and sports can provide genuinely unknown outcomes for new models.

\bibliographystyle{iclr2027_conference}
\bibliography{references}

\clearpage
\appendix

\section{Metric Formulae}
\label{app:formulas}


\noindent\textbf{Scoreline metrics.}
For a non-exact score prediction, the raw Scoreline value is

\begin{equation}
S_{\mathrm{score}}=45I_r+25\left[1-\frac{e_d}{5}\right]_+
+20\left[1-\frac{e_t}{6}\right]_+10\left[1-\frac{e_{team}}{8}\right]_+ .
\end{equation}

Here $I_r$ marks a correct result class, while $e_d$, $e_t$, and $e_{team}$ measure errors in goal difference, total goals, and team-wise goals. The 45/25/20/10 allocation was fixed before evaluation, and the error caps prevent large misses from producing negative scores.

The scoreline calibration uses center 70 and temperature 5. It preserves model order and the 0/100 endpoints. The composite calibration uses center 50 and temperature 5. Neither calibration is fitted to the evaluated systems. Individual layer displays use the same endpoint-preserving map with temperature 10.

If $\bar S_{\mathrm{score}}$ is the raw mean Scoreline value, its displayed value is

\begin{equation}
\widetilde S_{\mathrm{score}}=100\frac{\sigma((\bar S_{\mathrm{score}}-70)/5)-\sigma(-14)}{\sigma(6)-\sigma(-14)}.
\label{eq:scoreline-transform}
\end{equation}

\noindent\textbf{Availability-aware aggregation.} For any aggregation set $A$, the availability-aware weighted mean is

\begin{equation}
S_A=\frac{\sum_{j\in A}w_jS_j}{\sum_{j\in A}w_j}.
\end{equation}

At both aggregation levels, missing task values are omitted from the weighted mean. An observed absence is treated as a substantive outcome, so a confirmed match with no red card supports a correct prediction of no red card. Layer weights were specified before the experiments, and equal-weight sensitivity leaves the top three systems unchanged.

\noindent\textbf{Fixed contrast calibration.} The displayed composite uses

\begin{equation}
C(R)=100\frac{\sigma((R-50)/5)-\sigma(-10)}{\sigma(10)-\sigma(-10)},
\label{eq:composite-transform}
\end{equation}
where $\sigma(z)=1/(1+e^{-z})$.

\noindent\textbf{Competition-level evaluation.}
For competition evaluation, $G$ measures the agreement between predicted and actual group order using Kendall's $\tau$. $A_r$ is the fraction of correct teams reaching knockout round $r$, $M_r$ is the fraction of exact pairings in that round, and $H$ is champion accuracy. The round-level values are combined into weighted scores $A$ and $M$, with increasing weights from R32 to the final. A model therefore receives partial credit for selecting the right teams even when the bracket sides are wrong.
The completed competition score is

\begin{equation}
T5_{\mathrm{complete}}=\frac{0.25G+0.35(0.7A+0.3M)+0.20H}{0.80}.
\end{equation}

The competition-round weights are $1,2,4,8,16$ from R32 to the final. They are fixed before looking at model outputs and encode the increasing specificity of later-round claims.

\section{Reproducibility Protocol}

This appendix records the input contract and execution settings needed to reproduce the reported forecast artifacts. The paper's public release contains the frozen evidence snapshots, prompts, parsed predictions, truth records, and grading code. The fields below identify the information that must be retained when a new track is created.

\subsection{Fixed-evidence input}

The S1 input is a JSON object with a fixture header (competition, stage, kickoff time, venue, home team, and away team) and five evidence blocks: squad and availability information, recent results, team and player statistics, pre-lock news headlines with source metadata, and pre-lock market odds when available. The object also contains the UTC prediction deadline and the content hash of the frozen snapshot. No field is populated from a source published after that deadline. The S1 prompt asks for the common forecast object and a written rationale. It does not ask the model to call tools.

\subsection{Search-enabled input}

The S2 prompt contains the same fixture header and the same output instructions, but no curated evidence blocks. The agent may use the search tools exposed by its provider until the prediction deadline. Because providers expose different indexes and tool interfaces, we do not claim that S2 has a common search budget or that its difference from S1 is a retrieval-only effect. Each run records the provider, model identifier, tool configuration, query and response timestamps, retrieved URLs, tool-call count, latency, token counts, and the complete response. Sampling parameters (temperature, top-$p$, and output limit) use provider defaults unless an override is exposed. Any override is written to the run manifest. These records make the comparison auditable and state explicitly which settings are provider defaults.

\subsection{Output schema and repair}

Every pre-match response is a JSON object with the following top-level fields:
{
  \small
\begin{verbatim}

fixture, result_probabilities, predicted_result, predicted_score,
players, lineups, events, statistics,
competition_forecast, rationale
\end{verbatim}
}
The nested objects use fixed names for home/draw/away probabilities, score goals, player lists and rankings, event type/player/minute, and home/away statistic values. The parser checks JSON syntax, required keys, probability normalization, result--score consistency, and basic array types. If a response needs schema repair, the same prompt is sent at most once before the deadline. The original and repaired responses are both retained. No repair prompt contains post-match truth.

\subsection{Competition forecast timing}

The complete competition forecast is a single pre-tournament artifact. Group-order, bracket, finalist, and champion claims are elicited and frozen before the opening match. The program may derive later fixtures from the model's own group forecast, but it never asks the model to revise the bracket after an outcome is known. Post-tournament records enter only during grading.

\subsection{Metric audit}

The release stores both raw and displayed values for every task, layer, and composite. Task-to-layer and layer-to-composite aggregation uses the fixed weights in Table~\ref{tab:taxonomy}. Equation~\ref{eq:composite-transform} defines the endpoint-preserving display map, with $\sigma(z)=1/(1+e^{-z})$, center 50, and temperature 5. Layer plots use the same map with temperature 10. Recomputing the composite from the saved raw layer values therefore does not require any model call or hidden cohort normalization.

\subsection{Scope and transfer}

The World Cup track is a single empirical case study. It demonstrates the protocol under one competition format, but it cannot establish that the observed ranking or search behavior transfers unchanged to domestic leagues, less prominent competitions, or other sports. New tracks should register their own schedule, truth adapter, and competition rules while retaining the same pre-event lock, output schema, metric definitions, and audit fields. Such prospective tracks are the appropriate test of generalization.

\end{document}